\documentclass[letterpaper, 10 pt, conference]{ieeeconf}  

\IEEEoverridecommandlockouts                              
 
\usepackage{etoolbox}
\usepackage{balance}
\usepackage{booktabs}
\usepackage{cite}
\usepackage{subcaption}
\usepackage{graphicx}
\usepackage{multirow}
\usepackage[table]{xcolor}
\usepackage{pifont}
\newcommand{\cmark}{\ding{51}}
\newcommand{\xmark}{\ding{55}}
\usepackage{pifont}
\usepackage{hyperref}
\usepackage{cuted}   
 
\title{\LARGE \bf
SCD4VPR: Multi-modal Scene Change Detection for\\
Long-term Visual Place Recognition Database Update
}
 
\author{Diwei Sheng$^{1}$\textsuperscript{*}, Vijayraj Gohil$^{1}$\textsuperscript{*}, Satyam Gaba$^{1}$,
        Zihan Liu$^{1}$,\\ 
        Giles Hamilton-Fletcher$^{1}$, John-Ross Rizzo$^{1}$, Yongqing Liang$^{1}$, 
        and Chen Feng$^{1}$\textsuperscript{\ding{41}}
\thanks{* Equal contribution.}%
\thanks{\ding{41} Corresponding author
        ({\tt\small \href{mailto:cfeng@nyu.edu}{cfeng@nyu.edu}}).
        NYC-CD and code will be released upon publication.}%
\thanks{$^{1}$New York University}%
}
 
\IEEEaftertitletext{\vspace{-3\baselineskip}}

\begin{document}

\maketitle
\thispagestyle{empty}
\pagestyle{empty}
 
\begin{strip}
\centering
\begin{minipage}[t]{0.60\textwidth}
    \centering
    \includegraphics[width=\linewidth]{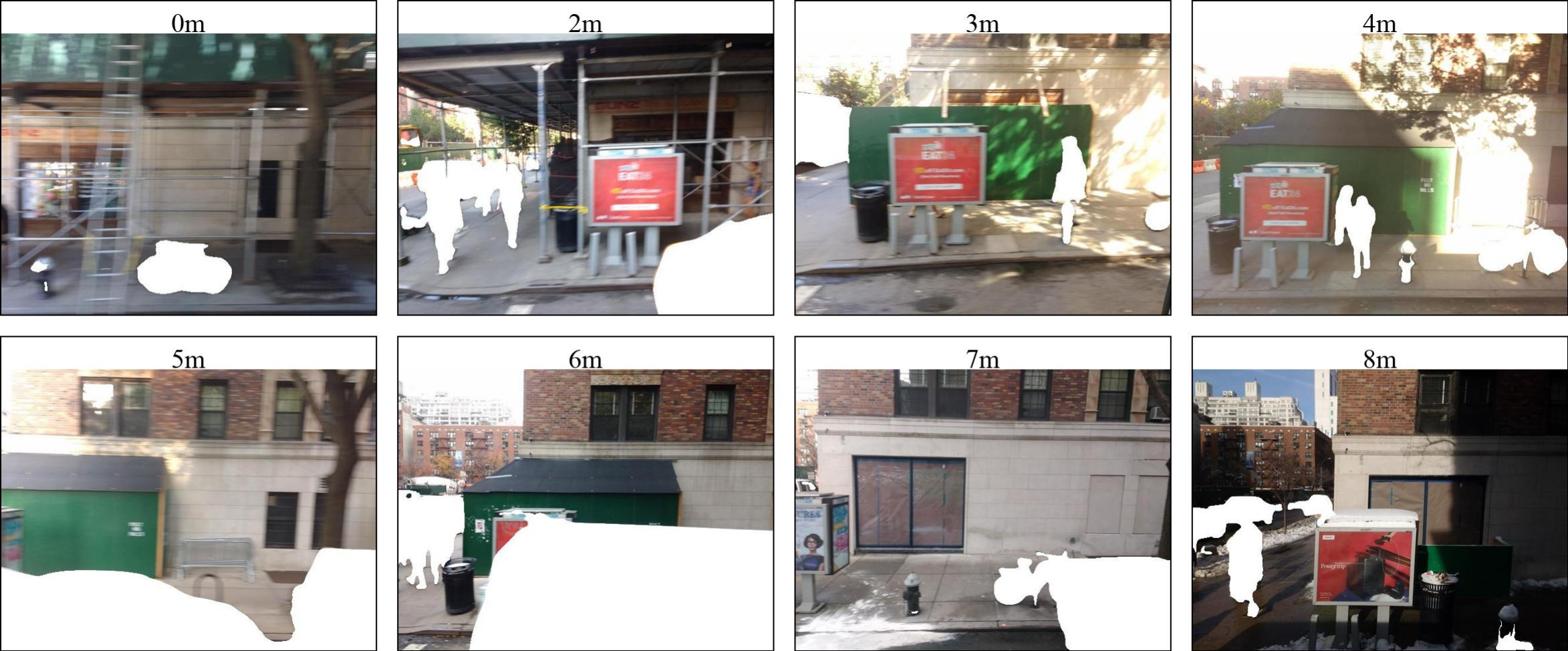}\\[2pt]
    (a)
\end{minipage}\hfill
\begin{minipage}[t]{0.37\textwidth}
    \centering
    \includegraphics[width=\linewidth]{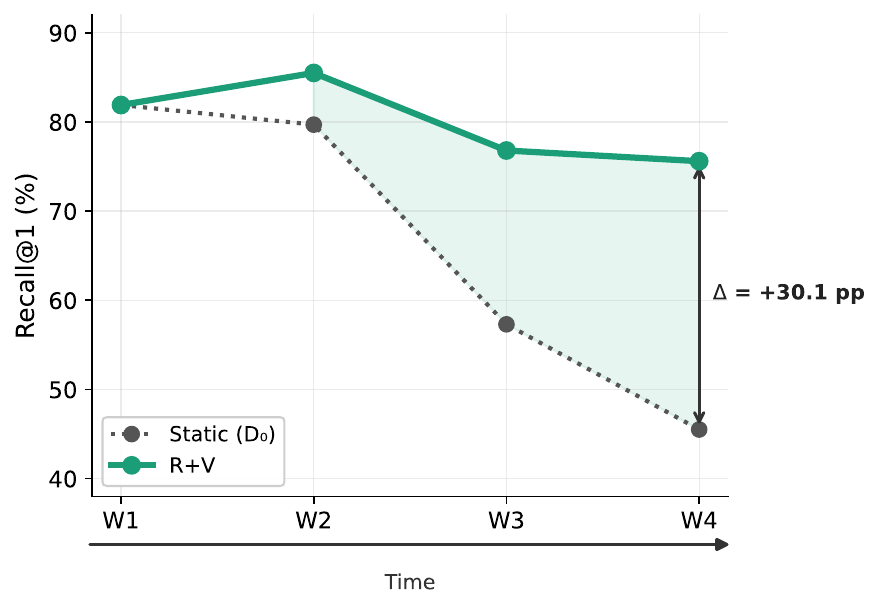}\\[2pt]
    (b)
\end{minipage}
\captionof{figure}{(a) Example place
         observed over 0--8 months in NYU-VPR, illustrating the
         kind of construction-driven appearance drift
         that static databases fail to track. (b) MixVPR~\cite{ali2023mixvpr} Recall@1 on side-view
         NYU-VPR~\cite{sheng2021nyu} across four temporal windows (W1--W4, summer
         through late winter). A static database
         ($D_0$) degrades sharply as seasonal appearance change
         accumulates, while SCD4VPR's R+V update strategy maintains
         recall, recovering $+30.1$ points at W4.}
\label{fig:vpr_deter}
\end{strip}

\begin{abstract}
 
Long-term autonomy in mobile robotics requires maps that remain accurate as environments change over time. Visual Place Recognition (VPR), a core localization capability, degrades sharply as the temporal gap between query and database images grows, particularly across seasonal transitions. Scene Change Detection (SCD) offers a principled mechanism for database maintenance, but existing methods rely on binary, uni-modal visual features that cannot distinguish structural changes from viewpoint-induced differences — a distinction essential for correct update decisions. We propose SCD4VPR, a scene change detection that jointly reasons about what has changed and distinguishes genuine change from viewpoint-induced difference in a unified vision-language framework. SCD4VPR fuses VLM-generated semantic descriptions with visual features via cross-modal attention and refines predictions with geometric-semantic matching, producing multi-class change masks that separately identify object changes, appearance changes, and viewpoint-induced changes. We introduce NYC-CD, the first real-world street-view SCD benchmark with pixel-level multi-class annotations across 8,122 image pairs. Experiments across four SCD benchmarks show that SCD4VPR consistently improves three architecturally distinct backbones. In a controlled VPR database maintenance experiment on NYU-VPR spanning summer through late winter, we confirm that retrieval performance deteriorates substantially when the database is left unchanged, and show that SCD4VPR-guided updates recover most of this loss (+30.1 R@1 at the largest time gap) while keeping the database far more compact than naïve append.
 
\end{abstract}

\section{INTRODUCTION}
\label{sec:intro}

Long-term autonomy in mobile robotics depends on maps that remain accurate
as environments change over time~\cite{pomerleau2014long, lowry2015visual}.
Visual Place Recognition (VPR)---retrieving the database image captured at the same place as a query image---is sensitive to this: as the temporal gap between query and database images grows, retrieval recall degrades sharply, especially across seasonal transitions such as summer to winter (Fig.~\ref{fig:vpr_deter}(b)). Fig.~\ref{fig:vpr_deter}(a) illustrates a concrete instance of this drift: a real-world Manhattan scene captured over several months, where ongoing construction progressively altered the environment. Static databases become obsolete as these changes accumulate, causing localization failures that compound over repeated deployments.
 
Scene Change Detection (SCD) offers a principled mechanism for keeping such
databases current~\cite{sousa2023systematic, cheng2025lt}. By identifying
which parts of a scene have changed, SCD enables selective updates:
incorporating new structures, pruning removed ones, and refreshing
appearance-altered entries. In contrast, a na\"ive strategy that appends every
newly captured image leads to unbounded database growth: accumulating
near-duplicates inflate the candidate pool, degrading retrieval precision and
increasing the risk of perceptual aliasing~\cite{lowry2015visual}. Realizing
the full benefit of SCD-guided maintenance requires correctly distinguishing
\textit{structural changes} (new objects, removed structures, seasonal
appearance shifts) from \textit{viewpoint-induced changes} (regions that fall outside the shared field of view due to shifts in camera pose). A method that cannot make this distinction will either miss genuine
updates or overwrite valid spatial coverage with shifted viewpoints---both
outcomes degrade VPR performance.
 
Despite significant progress, existing SCD methods handle this distinction
poorly. Supervised approaches based on CNNs~\cite{wang2023reduce,
alcantarilla2018street} or vision transformers~\cite{lin2025robust} rely on
low-level visual consistency, producing masks that are both fragmented and prone to mistaking shadows, reflections, and viewpoint shifts for structural changes. Recent pipelines
leveraging vision foundation models~\cite{kim2025towards} improve
generalization but still struggle to reason about \textit{what} changed in a
semantically meaningful way. A key missing ingredient is high-level semantic
reasoning: understanding not just that pixel intensities differ, but that a
construction scaffold appeared, that foliage turned bare, or that a facade visible from one viewpoint simply falls outside another's field of view. Recent work on Vision-Language model (VLMs) demonstrates strong semantic
reasoning capabilities~\cite{hurst2024gpt}, but directly applying off-the-shelf
VLMs to SCD is insufficient: VPR is often under large viewpoint shifts, VLMs frequently
produce hallucinated regions rather than precise change masks, and geometric
constraints must be enforced alongside linguistic
reasoning~\cite{deng2025visual}.
 
To address these limitations, we propose \textbf{SCD4VPR}, a vision-language
framework for semantic scene change detection that jointly reasons about
\textit{what} has changed and \textit{whether} an apparent difference is genuine or viewpoint-induced. Our
key insight is that language provides structured semantic priors describing
changed objects, while geometric reasoning ensures spatial consistency across
views. Crucially, SCD4VPR produces \textit{multi-class} change masks that
separately identify object changes, appearance changes, and viewpoint-induced
changes---enabling principled downstream decisions such as database
replacement, appearance refresh, and viewpoint coverage augmentation.
 
We validate SCD4VPR on two fronts. First, extensive experiments across four
street-view SCD benchmarks demonstrate that SCD4VPR consistently improves three
architecturally distinct backbones (CNN, transformer, and
foundation-model-based). Second, through a controlled map update
experiment on the NYU-VPR dataset~\cite{sheng2021nyu}, we show that
SCD4VPR's multi-class output enables update decisions that maintain
higher VPR recall over four temporal windows spanning summer through late
winter, while keeping database size substantially smaller than na\"ive append
strategies. The contributions of this paper are as follows:
\begin{itemize}
 
  \item We present \textbf{NYC-CD}, the first multi-class real-world
  street-view SCD benchmark, featuring pixel-level annotations for object
  changes, vegetation appearance changes, and viewpoint-induced changes in
  Manhattan, together with a semi-automatic annotation pipeline.
 
  \item We propose \textbf{SCD4VPR}, a modular framework that fuses
  VLM-generated semantic descriptions with visual features via cross-modal
  attention and refines predictions with geometric-semantic matching, producing
  precise, object-aware multi-class change masks robust to viewpoint variation
  and appearance shift.
 
  \item We demonstrate through a \textbf{VPR database maintenance experiment}
  that SCD4VPR-guided updates outperform na\"ive append and na\"ive replacement
  strategies in VPR recall over time, achieving the best
  recall-to-database-size efficiency across all evaluated strategies.
 
\end{itemize}

\section{Related Work}
\label{sec:related_works}

\subsection{Visual Place Recognition and Long-Term Map Maintenance}

VPR performance degrades as the gap between query and database images
widens, especially across seasonal transitions~\cite{sheng2021nyu},
as shown in Fig.~\ref{fig:vpr_deter}(b). Existing work addresses this
primarily through descriptor robustness---training models to be
invariant to appearance change~\cite{ali2023mixvpr,izquierdo2024optimal}---but this approach
does not prevent the database from becoming stale. Complementary directions
operate at the retrieval stage itself: language-augmented benchmarks and
methods such as LaVPR~\cite{idan2026lavpr} embed natural-language
descriptions directly into the retrieval representation, while
geometry-centric re-ranking methods such as VGGT-MPR~\cite{xu2026vggt}
refine multimodal candidate rankings after retrieval using cross-view
geometric consistency. Both are descriptor- or ranking-stage improvements
rather than maintenance-stage decisions about what the database should
contain, and are complementary to the update mechanism we propose.

Orthogonal to retrieval-stage improvements, explicit database maintenance
via map update~\cite{pomerleau2014long, sousa2023systematic, molloy2020intelligent,churchill2012practice}
addresses staleness directly. Na\"ive append
strategies grow the database without bound, degrading retrieval
efficiency~\cite{lowry2015visual,sousa2023systematic}. Recent work explores 3D-reconstruction-based
update~\cite{cheng2025lt} and multimodal LLM-based temporal change
analysis~\cite{deng2025visual}, but neither provides a principled mechanism
for distinguishing structural changes from viewpoint-induced differences at
the image level. We address this gap by using SCD4VPR's multi-class output to
guide database update decisions directly from image pairs, without requiring
3D reconstruction or GPS.

\subsection{Scene Change Detection Methods}

The majority of SCD methods rely on uni-modal visual information. Supervised
approaches using CNNs~\cite{wang2023reduce,
alcantarilla2018street} or vision
transformers~\cite{lin2025robust} achieve strong results in
controlled settings but, lacking semantic context for changed objects, produce masks that are fragmented and easily disturbed by shadows and reflections. Methods based on 3D Gaussian splatting~\cite{cheng2025lt}
handle geometric variations well but require dense multiview captures.
Weakly supervised~\cite{li2024semicd} and self-supervised~\cite{alpherts2025emplace}
approaches reduce annotation requirements but remain limited to visual
patterns. Training-free zero-shot methods such as
GeSCF~\cite{kim2025towards} offer generalization through pretrained
foundation models but are still constrained by visual-only features.

Language-integrated SCD is an emerging direction. Remote sensing
methods~\cite{dong2024changeclip, wang2025change} leverage language for
structured top-down changes such as building construction or deforestation,
but lack the geometric complexity of street-level viewpoint variation and
do not produce complete object-level masks. ViewDelta~\cite{varghese2024viewdelta}
requires user-provided text prompts, limiting practical deployment. Recent
work using multimodal LLMs for temporal street-level change
analysis~\cite{deng2025visual} demonstrates strong semantic understanding but
does not produce pixel-level change masks or distinguish change types. Our
SCD4VPR addresses these limitations by automatically generating change
descriptions via VLMs~\cite{hurst2024gpt}, fusing them with visual features
through cross-modal attention, and verifying predictions geometrically to
produce precise multi-class change masks.

\subsection{Scene Change Detection Datasets}

Table~\ref{tab_1} compares existing SCD datasets along three capabilities:
object change, appearance change, and viewpoint difference. Most prior datasets focus on object-level changes from aligned viewpoints~\cite{jst2015change, alcantarilla2018street, sakurada2020weakly, kim2025towards}, and even CityPulse~\cite{huang2024citypulse}, a related street-level time-series benchmark, provides only image-level
``change''/``no-change'' labels rather than pixel-level masks; none jointly support all three change types in real street-view environments. Simulated datasets~\cite{park2021changesim, mata2022standardsim} provide scale but do not capture real-world urban complexity. Our NYC-CD dataset fills this gap with 8,122 real-world image pairs annotated with pixel-level masks for object changes, vegetation appearance changes, and viewpoint-induced changes in Manhattan.

\begin{table}[t]
\vspace{4pt}
\caption{Comparison of major SCD datasets with our NYC-CD dataset. \textbf{Obj.}: object change, \textbf{App.}: appearance change, \textbf{VP.}: viewpoint difference.}
\centering
\setlength{\tabcolsep}{4pt}
\renewcommand{\arraystretch}{0.92}
\resizebox{\columnwidth}{!}{%
\begin{tabular}{l c c c c c}
\toprule
\textbf{Dataset} & \textbf{Real/Sim} & \textbf{Pairs}
& \textbf{Obj.} & \textbf{App.} & \textbf{VP.} \\
\midrule
CARLA-OBJCD~\cite{hamaguchi2020epipolar} & Sim   & 15000        & \cmark & \xmark & \cmark \\
Changesim~\cite{park2021changesim}        & Sim   & $\sim$130k   & \cmark & \cmark & \xmark \\
Standardsim~\cite{mata2022standardsim}   & Sim   & 12718        & \cmark & \xmark & \xmark \\
\midrule
CD2014~\cite{wang2014cdnet}              & Real  & 7000         & \cmark & \xmark & \xmark \\
PCD~\cite{jst2015change}                 & Real  & 200          & \cmark & \xmark & \xmark \\
VL-CMU-CD~\cite{alcantarilla2018street}  & Real  & 1362         & \cmark & \xmark & \xmark \\
PSCD~\cite{sakurada2020weakly}           & Real  & 770          & \cmark & \xmark & \xmark \\
GSV-OBJCD~\cite{hamaguchi2020epipolar}   & Real  & 500          & \cmark & \xmark & \cmark \\
UMAD~\cite{li2024umad}                   & Real  & 26301        & \cmark & \xmark & \xmark \\
ChangeVPR~\cite{kim2025towards}          & Real  & 529          & \cmark & \xmark & \xmark \\
CityPulse~\cite{huang2024citypulse}      & Real  & 25423        & \cmark & \xmark & \xmark \\
\midrule
\rowcolor[HTML]{F2EDF9}
\textbf{NYC-CD (Ours)}                   & \textbf{Real} & \textbf{8122}
& \textbf{\cmark} & \textbf{\cmark} & \textbf{\cmark} \\
\bottomrule
\end{tabular}
}
\label{tab_1}
\vspace{-2mm}
\end{table}

\begin{figure}[t]
    \centering
    \includegraphics[width=\columnwidth]{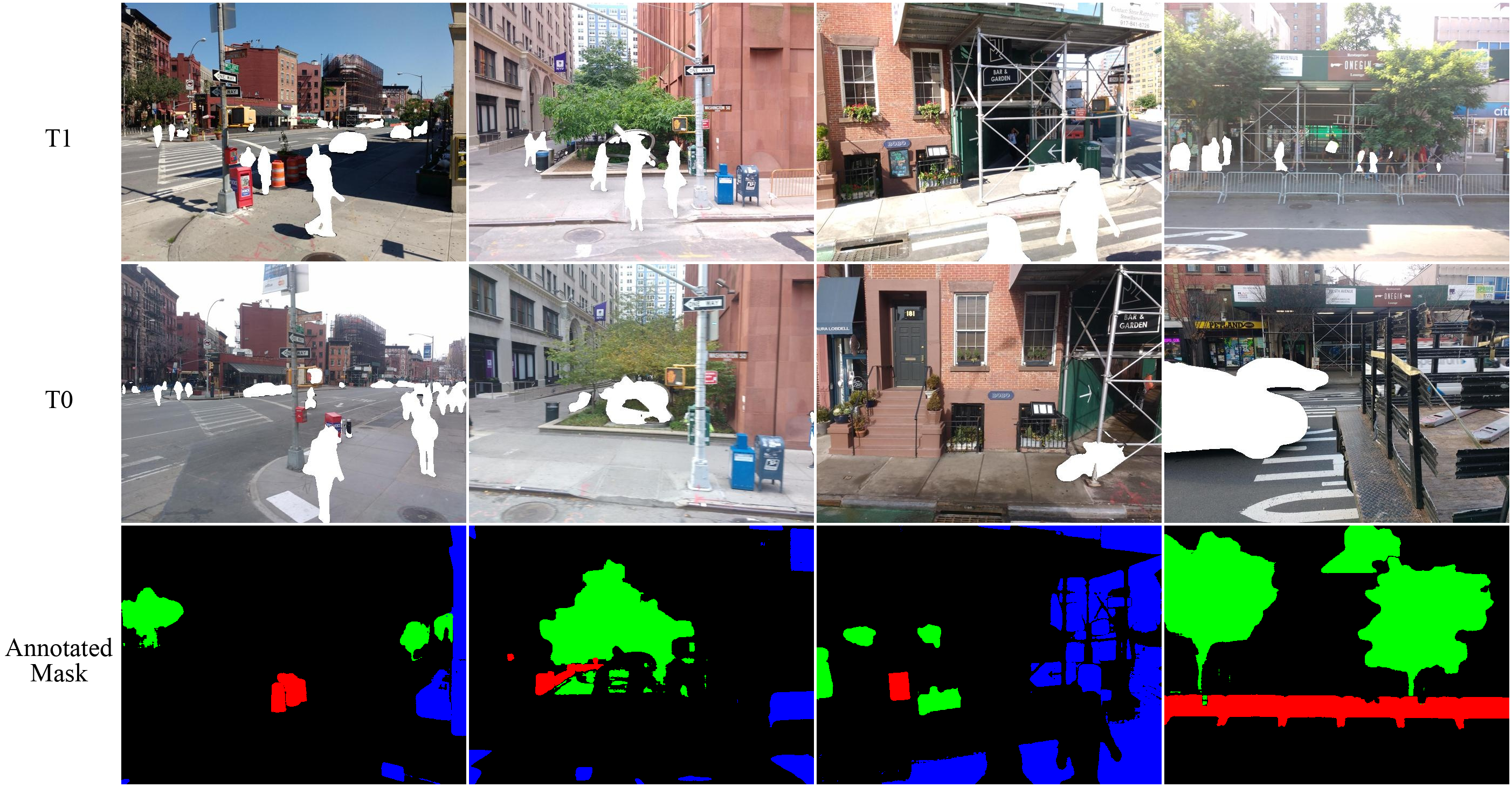}
    \caption{Four image pairs annotated by our pipeline. Each pair shows
    $T_0$ (top) and $T_1$ (bottom) with multi-class change masks:
    \textcolor{red}{new objects},
    \textcolor{green!60!black}{appearance changes}, and
    \textcolor{blue}{viewpoint-induced changes}.}
    \label{fig:anony}
    \vspace{-2mm}
\end{figure}

\section{The NYC-CD Dataset}

NYC-CD is a large-scale real-world SCD benchmark comprising 8,122
image pairs annotated with pixel-level masks for three change types
on $T_1$ with respect to $T_0$: \textbf{(1) new/missing objects};
\textbf{(2) vegetation appearance changes}; and \textbf{(3) viewpoint-induced changes} (regions
outside the shared field of view due to camera pose shift). This
multi-class taxonomy is the first in a real-world street-view SCD
benchmark, enabling downstream systems to act on each change type rather than treating all changes as equivalent.

\subsection{Image Pair Construction}

NYC-CD is extracted from the NYU-VPR dataset~\cite{sheng2021nyu},
which contains street-level imagery recorded throughout Manhattan
from May 2016 to March 2017. All images carry GPS tags for spatial
reference; pedestrians and vehicles are anonymized
using MSeg~\cite{MSeg_2020_CVPR} to preserve privacy.

We form image pairs using a two-stage approach: GPS proximity ensures geographic co-location, while MixVPR~\cite{ali2023mixvpr} retrieval selects the most visually similar candidate within that geographic neighborhood, maintaining sufficient visual overlap for change detection. Images from different quarters are used as database and queries (Q1 paired with Q3, Q2 paired with Q4), enforcing a minimum three-month temporal gap that ensures meaningful scene changes are captured. 

\subsection{Semi-Automatic Annotation Pipeline}

Given a pair ($T_0$, $T_1$), a VLM is prompted separately for object and
vegetation changes, and the resulting descriptions are grounded into
pixel-level semantic masks by Grounded SAM~\cite{ren2024grounded}. In
parallel, SAM2~\cite{ravi2024sam} tracks class-agnostic segments from $T_1$
to $T_0$ and flags those that are geometrically inconsistent---present in $T_1$ but not in $T_0$. We retain only segments where
the two cues agree, filtering out shadows, reflections, and other
distractors.
MAST3R~\cite{leroy2024grounding} then estimates dense correspondences to
determine the common-view region: retained proposals inside it are labeled
object or appearance changes, while those outside are labeled viewpoint-induced changes, reflecting field-of-view loss
from camera pose shift. A final manual pass discards pairs with missed changes and removes false
positive object masks.
Among the 8,122 curated pairs, 5,040 contain object changes, 5,093 contain
vegetation appearance changes, and 7,201 exhibit viewpoint-induced changes;
categories are non-exclusive. Fig.~\ref{fig:anony} shows example
annotations.


\section{SCD4VPR Methodology}
\label{sec:method}

\subsection{Overall Architecture}

\begin{figure*}[t]
  \vspace{6pt}
  \centering
  \includegraphics[width=0.8\textwidth]{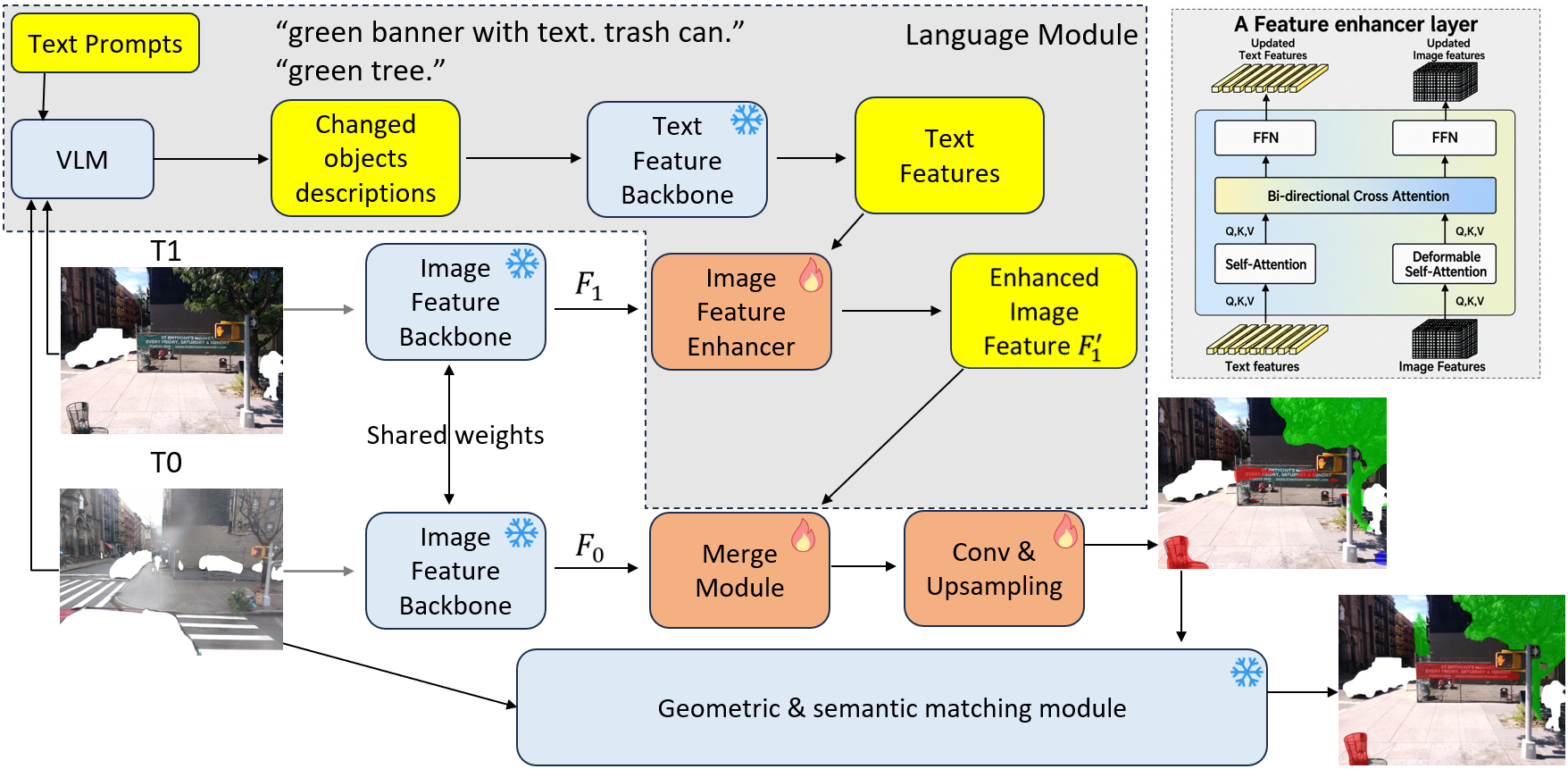}
  \vspace{-1mm}
  \caption{Overview of the proposed SCD4VPR. Two images ($T_0$, $T_1$)
  are processed by a shared-weight image encoder. The Cross-Modal
  Feature Enhancer injects language-derived priors about new objects
  in $T_1$ into visual features, producing an initial change mask.
  The Geometric--Semantic Matching Module refines this mask by
  enforcing spatial completeness and semantic consistency.}
  \label{fig:pipeline}
\end{figure*}

SCD4VPR augments standard SCD backbones with two plug-and-play modules,
as illustrated in Fig.~\ref{fig:pipeline}. A standard SCD backbone
consists of a shared-weight image encoder, a feature merging module,
and a segmentation head; existing methods differ primarily in whether
the encoder is CNN-based, DINOv2-based, or
SAM-based~\cite{wang2023reduce, lin2025robust, kim2025towards}.
The \textbf{Cross-Modal Feature Enhancer} takes the encoded feature
of $T_1$ and enhances it with text features describing scene changes,
producing an updated representation $F_1'$. This enhanced feature is
then compared with $T_0$ features through the backbone's merging module
and segmentation head to produce an initial pixel-level change mask.
The \textbf{Geometric--Semantic Matching Module} subsequently refines
this mask into object-consistent predictions by jointly leveraging
SAM2's class-agnostic tracking and Grounded SAM's semantic
segmentation. Together, these modules integrate semantic reasoning
and geometric consistency while remaining compatible with any SCD
backbone.

\subsection{Language Module and Cross-Modal Enhancer}

Given an image pair ($T_0$, $T_1$), a VLM 
generates a natural language description of objects that appear in
$T_1$ but not in $T_0$, and objects with different appearances. This
description serves as a high-level hypothesis of potential scene
changes. The generated text is encoded by a frozen text feature backbone; we extract the full sequence of
token embeddings from the final hidden layer to obtain text features.

To integrate text features with visual features $F_1$, we employ a
cross-modal feature enhancer whose architecture adapts the feature
enhancement layers of GroundingDINO~\cite{liu2024grounding} to the
SCD setting (Fig.~\ref{fig:pipeline}, right). Specifically, we retain the bidirectional per-token
cross-attention design---where each image token attends over all text
tokens (text-to-image) and each text token refines its embedding from
visual context (image-to-text)---while replacing GroundingDINO's
detection-oriented image encoder with the SCD backbone's shared-weight
encoder. Deformable self-attention over image tokens enables adaptive
spatial focus, while text token self-attention preserves linguistic
structure. The enhanced features $F_1'$ ground visual representations
in semantic object descriptions, reducing false positives from
illumination variation, shadows, and reflections that mimic
structural changes.

\subsection{Geometric and Semantic Matching Module}

The geometric and semantic matching module refines the initial mask
from pixel-level responses to object-level reasoning by addressing
two key challenges: incomplete segmentation and noise from irrelevant
changes.

The \textbf{semantic matching} component employs Grounded
SAM~\cite{ren2024grounded} to validate semantic relevance and complete
the initial mask to full object extent. Because the initial mask is
produced at the pixel level, it can fragment a single object across
internal boundaries---e.g., splitting a car into separate regions at
its wheels---whereas Grounded SAM's object-level prompting yields a
single, complete mask per candidate object. Guided by the VLM-generated
descriptions, Grounded SAM produces masks aligned with candidate
changed objects; those satisfying overlap ratio $\alpha > \alpha_g$
with the initial mask are retained, filtering out false positives
caused by shadows, reflections, and other non-structural variations.

The \textbf{geometric matching} component leverages SAM2's tracking
capabilities~\cite{ravi2024sam} to isolate which instances
changed. Segment proposals are generated in $T_1$ and tracked
to $T_0$; since a semantic category (e.g., ``chair'') may correspond
to multiple instances, only those that are geometrically inconsistent
across views---absent or displaced in $T_0$---are retained. These
retained segments must satisfy overlap ratio $\alpha > \alpha_t$
with the initial mask, disambiguating the changed instance from
unchanged instances of the same category.

Both thresholds are calibrated on the NYC-CD validation set using the
RSCD+SCD4VPR configuration: a threshold sweep yields $\alpha_t = 0.20$
and $\alpha_g = 0.10$, which are then applied uniformly across all
backbones and datasets without any test-set tuning. Semantic and
geometric matching operate in synergy: semantic matching filters
noise and completes object masks, while geometric matching
disambiguates which instances of a given category changed,
together transforming noisy pixel-level predictions into coherent,
object-aware change masks. The resulting per-class labels (object,
appearance, viewpoint, no-change) drive downstream maintenance
decisions, as we demonstrate in Sec.~\ref{sec:vpr_maintenance}.

\begin{table*}[t]
\vspace{4pt}
\caption{F1 and IoU comparison of GeSCF, RSCD, and C-3PO with and without
SCD4VPR across four datasets. Bold: highest per dataset.
\colorbox{green!20}{Green}: improvement over baseline.}
\label{tab:model_comparison}
\centering
\setlength{\tabcolsep}{3pt}
\renewcommand{\arraystretch}{0.95}
\footnotesize
\begin{tabular}{lcccccccc}
\toprule
\textbf{Model}
& \multicolumn{2}{c}{\textbf{ChangeVPR}}
& \multicolumn{2}{c}{\textbf{VL-CMU-CD}}
& \multicolumn{2}{c}{\textbf{PSCD}}
& \multicolumn{2}{c}{\textbf{NYC-CD (ours)}} \\
\cmidrule(lr){2-3}\cmidrule(lr){4-5}\cmidrule(lr){6-7}\cmidrule(lr){8-9}
& F1 & IoU & F1 & IoU & F1 & IoU & F1 & IoU \\
\midrule
GeSCF          & 0.47 & 0.36 & 0.77 & 0.65 & 0.40 & 0.28 & 0.16 & 0.10 \\
GeSCF+SCD4VPR   & \cellcolor{green!20}\textbf{0.64} (+17\%) & \cellcolor{green!20}\textbf{0.55} (+19\%) & \cellcolor{green!20}0.82 (+5\%) & \cellcolor{green!20}0.71 (+6\%) & \cellcolor{green!20}0.49 (+9\%) & \cellcolor{green!20}0.37 (+9\%) & \cellcolor{green!20}0.57 (+41\%) & \cellcolor{green!20}0.49 (+38\%) \\
RSCD           & 0.24 & 0.16 & 0.83 & 0.73 & 0.54 & 0.40 & 0.13 & 0.17 \\
RSCD+SCD4VPR    & \cellcolor{green!20}0.63 (+39\%) & \cellcolor{green!20}0.54 (+38\%) & \cellcolor{green!20}\textbf{0.85} (+2\%) & \cellcolor{green!20}\textbf{0.75} (+2\%) & \cellcolor{green!20}0.59 (+5\%) & \cellcolor{green!20}0.46 (+6\%) & \cellcolor{green!20}\textbf{0.70} (+57\%) & \cellcolor{green!20}\textbf{0.58} (+51\%) \\
C-3PO          & 0.12 & 0.08 & 0.50 & 0.39 & 0.66 & 0.54 & 0.09 & 0.10 \\
C-3PO+SCD4VPR   & \cellcolor{green!20}0.58 (+46\%) & \cellcolor{green!20}0.49 (+41\%) & \cellcolor{green!20}0.73 (+23\%) & \cellcolor{green!20}0.67 (+28\%) & \cellcolor{green!20}\textbf{0.70} (+4\%) & \cellcolor{green!20}\textbf{0.58} (+4\%) & \cellcolor{green!20}0.57 (+48\%) & \cellcolor{green!20}0.47 (+37\%) \\
\bottomrule
\end{tabular}
\vspace{-2mm}
\end{table*}

\begin{table}[t]
\centering
\caption{Multi-class change detection on NYC-CD (RSCD backbone).}
\label{tab:multiclass_change}
\setlength{\tabcolsep}{3pt}
\renewcommand{\arraystretch}{0.92}
\resizebox{\columnwidth}{!}{%
\begin{tabular}{lcccc}
\toprule
\multirow{2}{*}{\textbf{Class}} & \multicolumn{2}{c}{\textbf{F1}} & \multicolumn{2}{c}{\textbf{IoU}} \\
\cmidrule(lr){2-3}\cmidrule(lr){4-5}
& RSCD & +SCD4VPR ($\Delta$) & RSCD & +SCD4VPR ($\Delta$) \\
\midrule
Non-change      & 0.899 & \cellcolor{green!20}\textbf{0.945} (+4.6\%) & 0.816 & \cellcolor{green!20}\textbf{0.895} (+7.9\%) \\
New/Missing Obj.& 0.438 & \cellcolor{green!20}\textbf{0.472} (+3.4\%) & 0.281 & \cellcolor{green!20}\textbf{0.309} (+2.8\%) \\
Appearance      & 0.655 & \cellcolor{green!20}\textbf{0.703} (+4.8\%) & 0.487 & \cellcolor{green!20}\textbf{0.542} (+5.5\%) \\
Viewpoint       & 0.459 & \cellcolor{green!20}\textbf{0.563} (+10.4\%) & 0.298 & \cellcolor{green!20}\textbf{0.391} (+9.3\%) \\
\midrule
\textbf{Mean}   & 0.613 & \cellcolor{green!20}\textbf{0.671} (+5.8\%) & 0.471 & \cellcolor{green!20}\textbf{0.534} (+6.3\%) \\
\bottomrule
\end{tabular}}
\vspace{-2mm}
\end{table}

\begin{table}[t]
\centering
\caption{Ablation on NYC-CD using RSCD as the base architecture.}
\label{tab:ablation}
\setlength{\tabcolsep}{4pt}
\renewcommand{\arraystretch}{0.92}
\footnotesize
\begin{tabular}{lcc}
\toprule
\textbf{Method} & \textbf{F1} & \textbf{IoU} \\
\midrule
RSCD                    & 0.127 & 0.168 \\
RSCD + Text (InternVL)  & 0.513 & 0.345 \\
RSCD + Text (QwenVL)    & 0.537 & 0.401 \\
RSCD + Text (GPT-4o)    & 0.563 & 0.424 \\
RSCD + Text + SM        & 0.584 & 0.487 \\
RSCD + Text + SM + GM   & \textbf{0.696} & \textbf{0.582} \\
\bottomrule
\end{tabular}
\vspace{-3mm}
\end{table}

\section{Experiments}
\label{sec:results}

We validate SCD4VPR in two stages: first as a general-purpose SCD
improvement across four benchmarks (Sec.~\ref{sec:scd_results}), then in
its motivating downstream application, VPR database maintenance
(Sec.~\ref{sec:vpr_maintenance}).

\subsection{SCD Benchmark Evaluation}
\label{sec:scd_results}

\subsubsection{Setup}

We plug SCD4VPR into three representative SCD backbones: C-3PO~\cite{wang2023reduce}
(CNN-based), RSCD~\cite{lin2025robust} (DINOv2-based transformer), and
GeSCF~\cite{kim2025towards} (SAM-based, zero-shot). Evaluation uses four
street-view datasets: NYC-CD, VL-CMU-CD~\cite{alcantarilla2018street},
PSCD~\cite{sakurada2020weakly}, and ChangeVPR~\cite{kim2025towards}, spanning
urban development, seasonal variation, and natural disasters. NYC-CD is split
into 5,195 / 1,299 / 1,630 train/val/test pairs. For binary evaluation, masks
are formed as the union of all change classes. For multi-class evaluation, RSCD
is extended with a four-class segmentation head trained on NYC-CD. As noted in
prior work~\cite{wang2023reduce, kim2025towards}, VL-CMU-CD contains evidently
incorrect annotations; we remove these pairs and use the filtered set for all
baselines to ensure fair comparison.

\textbf{VLM prompting and generation settings.} Change descriptions are
generated using GPT-4o with temperature 0.2, max tokens 4096, and default
top-\emph{p} (1.0); the API does not expose a stable seed, so runs are
near-deterministic under low temperature rather than exactly reproducible.
Each call receives both $T_0$ and $T_1$ as base64-encoded images alongside
a task-specific prompt. The vegetation-change prompt reads: \textit{``I have
2 images of outdoor scenes: A and B. ... give me a numbered list of changed
plant names in A, and a numbered list of changed plant names in B.''} The
object-change prompt follows the same two-image, dual-direction structure,
additionally instructing the model to exclude weather, lighting, vehicles,
people, and animals from its answer. The same prompting protocol is used for
both dataset annotation and inference, tested against InternVL and QwenVL in
Table~\ref{tab:ablation}.

\begin{figure*}[t]
    \centering
    \includegraphics[width=\textwidth]{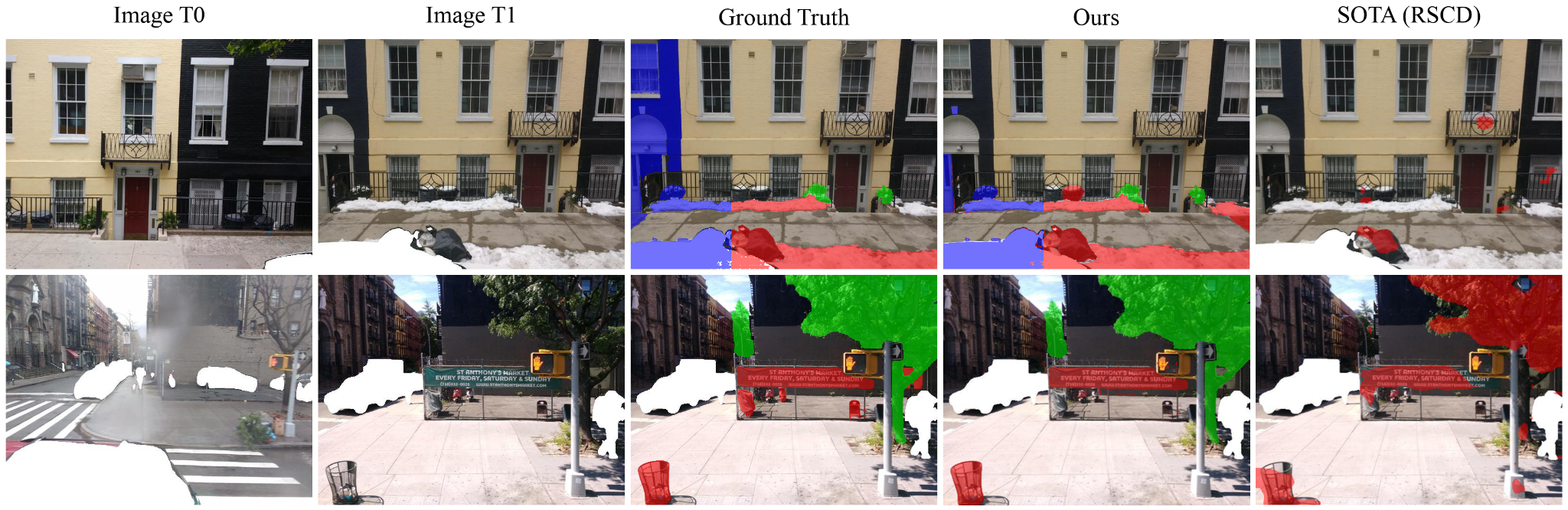}
    \caption{Qualitative comparison on NYC-CD. SCD4VPR produces
    semantically accurate multi-class masks distinguishing
    \textcolor{red}{object changes},
    \textcolor{green!60!black}{appearance changes}, and
    \textcolor{blue}{viewpoint-induced changes}, while RSCD
    (SOTA baseline) produces fragmented binary detections that
    miss or misclassify meaningful scene changes.}
    \label{fig:qualitative_result}
    \vspace{-3mm}
\end{figure*}

\subsubsection{Comparative Results}

Table~\ref{tab:model_comparison} reports F1 and IoU results across
all four datasets. SCD4VPR consistently improves all three baselines
across all datasets. Gains are largest on ChangeVPR (strong
appearance variation typical of long-term VPR): GeSCF +17\%, RSCD
+39\%, C-3PO +46\% F1. Improvements are more moderate but consistent
on VL-CMU-CD and PSCD. The largest gains appear on NYC-CD, our most
challenging benchmark due to viewpoint variation and multi-class
change complexity: RSCD+SCD4VPR achieves 0.70 F1 / 0.58 IoU
(+57\% / +51\%).

Multi-class results in Table~\ref{tab:multiclass_change} show gains
across all four change categories, with the largest improvement in
the viewpoint-induced class (+10.4\% F1), confirming that
language-guided reasoning and geometric matching are most beneficial
where visual methods struggle.
Fig.~\ref{fig:qualitative_result} confirms gains qualitatively:
SCD4VPR produces object-complete multi-class masks while RSCD yields
fragmented binary detections that miss or misclassify meaningful
changes.

\subsubsection{Ablation Study}

Table~\ref{tab:ablation} isolates the contribution of each component using RSCD
as the base. Language guidance alone yields substantial gains regardless of the
VLM used, confirming that the method is not tied to a specific proprietary
model and that improvements do not depend on using the same VLM for annotation
and inference. Semantic matching (SM) and geometric matching (GM) modules each
add gains, with the full system achieving the best performance.

The complete pipeline processes each image pair in 4.89\,s on average, dominated
by caption generation (86\%). For the offline map maintenance use case here, reliability of change identification is more critical than
per-frame speed.

\subsection{VPR Database Maintenance}
\label{sec:vpr_maintenance}

\subsubsection{Experimental Setup}

We evaluate the utility of RSCD~\cite{lin2025robust} augmented with our
SCD4VPR modules
trained on NYC-CD, for long-term VPR database
maintenance using the NYU-VPR dataset~\cite{sheng2021nyu}; NYC-CD images
are excluded from the maintenance study
to avoid train/test overlap. An initial database
$D_0$ is constructed from side-view images captured April--August
2016 (20,692 images). Four subsequent temporal windows simulate
ongoing robot operation: W1 (Sep--Oct 2016, late summer$\to$fall),
W2 (Nov--Dec 2016, fall$\to$early winter), W3 (Jan 2017, peak
winter), and W4 (Feb--Mar 2017, late winter$\to$early spring).

For each window, the newly captured images serve
as evaluation queries and as update candidates, mirroring real
deployment where a robot's traversal provides the localization
queries and the maintenance signal. Images are retained if they have
a GPS-matched database entry within 25\,m, then further restricted
to locations covered in the previous window, ensuring
consistent spatial coverage across windows. The resulting sets are spatially subsampled
at 1\,m to remove near-duplicate traversals, yielding 7,257 / 5,905
/ 2,694 / 2,446 queries per window. Evaluation precedes update at
each window: Recall@1 is measured using MixVPR~\cite{ali2023mixvpr} against
the current database, then the database is updated using that
window's images for the next evaluation cycle. A retrieval is considered
successful if the returned image lies within 25\,m of the query.

For each image $N$, the top-1 MixVPR match $D^*$ from the
evaluation step is passed to MAST3R dense
matching~\cite{leroy2024grounding} to verify geometric consistency
(inlier ratio $\geq 0.3$); pairs below threshold are skipped to
reject perceptual aliases. SCD4VPR then classifies the change type
of the verified pair to determine the update action.

\textbf{Sensitivity of the geometric verification threshold.} We evaluate
threshold sensitivity on 9{,}934 MixVPR top-1 retrieval pairs from a side-view
NYU-VPR update window, treating pairs within 25\,m as genuine same-location matches and pairs
beyond 200\,m as far-field, non-corresponding retrievals. Inlier ratio degrades
gradually with distance (genuine mean $0.96$; far-field mean $0.49$). At $\tau=0.3$, 99.9\% of genuine pairs are retained
while 86.3\% of far-field pairs are rejected (precision $0.93$); we adopt
this threshold to prioritize not discarding genuine updates.

\subsubsection{Update Strategies}

Five strategies are compared in Table~\ref{tab:update_strategies},
each mapping SCD4VPR's per-pair output $(R, G, B, K)$ to an action
in \{replace, add, skip\}. \textbf{NU} ignores SCD4VPR's output
entirely, replacing $D^*$ for every geometrically verified pair.
\textbf{R} uses SCD4VPR to replace only on structural changes,
skipping pairs where only viewpoint-induced change is detected.
\textbf{R+V} extends R by \textit{adding} the new image when
viewpoint-induced change is detected, augmenting angular coverage
rather than overwriting. \textbf{A} adds all detected changes without
replacement. \textbf{NA} appends all geometrically verified images, requiring no SCD inference.

\begin{table}[t]
\centering
\caption{Database update strategies. Each column shows the action
taken for a given SCD4VPR per-pair output category---object change
(\textbf{Obj.}), appearance change (\textbf{App.}), viewpoint-induced
change (\textbf{VP.}), or no change (\textbf{No-chg.}).}
\label{tab:update_strategies}
\setlength{\tabcolsep}{4pt}
\renewcommand{\arraystretch}{1.1}
\resizebox{\columnwidth}{!}{%
\begin{tabular}{lccccc}
\toprule
\textbf{Strategy} & \textbf{Obj.} & \textbf{App.} & \textbf{VP.} & \textbf{No-chg.} & \textbf{DB size} \\
\midrule
NU (na\"ive update) & replace & replace & replace & replace & constant \\
R (replace)         & replace & replace & skip    & skip    & constant \\
R+V (replace+viewpoint-add) & replace & replace & \textbf{add} & skip & grows \\
A (add-all)         & add     & add     & add     & skip    & grows \\
NA (na\"ive append) & add     & add     & add     & add     & grows \\
\bottomrule
\end{tabular}}
\vspace{-2mm}
\end{table}

\subsubsection{Results}

\begin{figure}[t]
    \vspace{6pt}
    \centering
    \includegraphics[width=0.9\columnwidth]{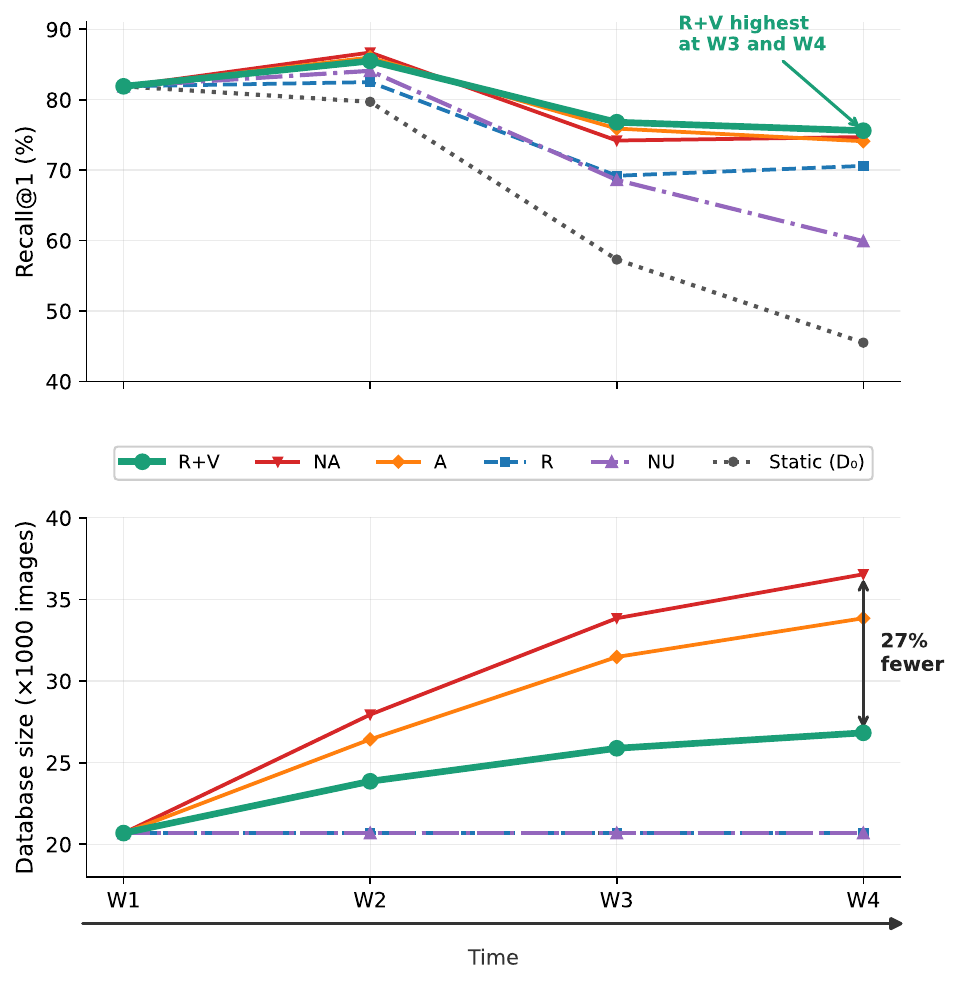}
    \caption{(\textit{Top}) Recall@1 over four temporal windows; all
    strategies evaluate against the current database before updating.
    (\textit{Bottom}) Database size over time. R+V achieves the highest
    recall at W3 and W4 while maintaining a 27\% smaller database than NA. NU, R, and $D_0$ maintain constant.}
    \label{fig:vpr_results}
    \vspace{-3mm}
\end{figure}

\begin{figure}[t]
    \vspace{6pt}
    \centering
    \includegraphics[width=0.8\columnwidth]{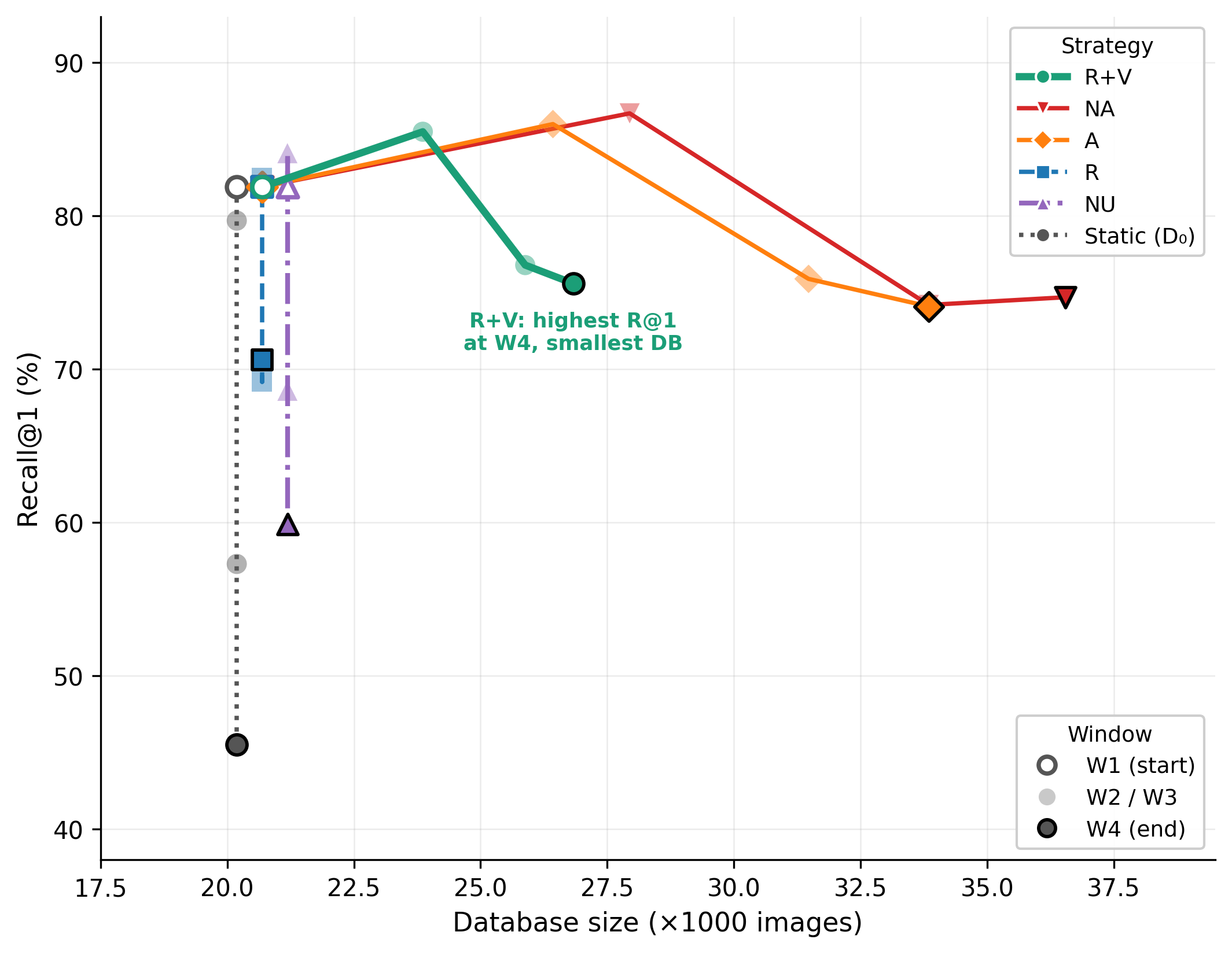}
    \caption{Recall@1 vs.\ database size across four update windows. Marker fill shows progression from Window~1 (hollow) to Window~4
(solid, outlined); Windows~2--3 are faded. Static (D$_0$), R, and NU maintain
a constant database size of 20,687 images across all windows; their markers
are offset horizontally by $\pm$500 images purely for visual separation and
do not reflect actual database size differences.}
    \label{fig:vpr_scatter}
\end{figure}

\begin{table}[t]
\caption{Extended recall metrics (R@1/5/10/20) by strategy and window.}
\centering
\setlength{\tabcolsep}{3pt}
\renewcommand{\arraystretch}{0.8}
\resizebox{\columnwidth}{!}{%
\begin{tabular}{c l c c c c}
\toprule
\textbf{Window} & \textbf{Strategy} & \textbf{R@1} & \textbf{R@5} & \textbf{R@10} & \textbf{R@20} \\
\midrule
\multirow{6}{*}{W2}
 & Static ($D_0$) & 79.7 & 86.6 & 88.0 & 89.5 \\
 & NU             & 84.1 & 90.9 & 92.0 & 93.0 \\
 & R              & 82.5 & 89.5 & 90.9 & 92.4 \\
 & A              & 86.0 & 92.0 & 93.2 & 94.1 \\
 & NA             & \textbf{86.7} & \textbf{92.4} & \textbf{93.6} & \textbf{94.4} \\
 & R+V            & 85.5 & 91.8 & 93.0 & 94.1 \\
\midrule
\multirow{6}{*}{W3}
 & Static ($D_0$) & 57.3 & 70.9 & 74.7 & 78.1 \\
 & NU             & 68.6 & 81.3 & 83.5 & 85.9 \\
 & R              & 69.2 & 79.8 & 81.7 & 83.7 \\
 & A              & 75.9 & 81.8 & 83.6 & 86.0 \\
 & NA             & 74.2 & 82.0 & 83.7 & 85.9 \\
 & R+V            & \textbf{76.8} & \textbf{82.5} & \textbf{84.9} & \textbf{86.3} \\
\midrule
\multirow{6}{*}{W4}
 & Static ($D_0$) & 45.5 & 55.9 & 62.7 & 75.0 \\
 & NU             & 59.9 & 78.6 & 81.9 & 83.6 \\
 & R              & 70.6 & 76.9 & 79.1 & 80.6 \\
 & A              & 74.1 & 83.7 & 87.2 & 89.7 \\
 & NA             & 74.7 & 84.0 & 87.4 & 90.2 \\
 & R+V            & \textbf{75.6} & \textbf{84.5} & \textbf{87.8} & \textbf{90.8} \\
\bottomrule
\end{tabular}
}
\label{tab:extended_recall}
\vspace{-4mm}
\end{table}
Fig.~\ref{fig:vpr_results} shows R@1 over time and database size over time
for all five strategies. Fig.~\ref{fig:vpr_scatter} shows the recall
vs.\ database size tradeoff at Window 4.

\textbf{R+V leads from Window 3 onward.} R+V achieves the highest R@1 at
Windows 3 and 4 (peak and late winter), precisely when the static database
$D_0$ is most stale. At Window 4, all update strategies substantially
outperform the static baseline, confirming that database maintenance is
essential for long-term VPR performance.

\textbf{NU progressively degrades at R@1.} Despite maintaining a
constant-size database, NU's R@1 declines steadily across windows (W2:
84.1 $\to$ W3: 68.6 $\to$ W4: 59.9). Blind replacement, including pairs
where only viewpoint-induced change is detected, overwrites valid
spatial coverage with shifted-viewpoint images, cumulatively damaging
rank-1 retrieval. R+V avoids this by preserving the original entry and
appending the new viewpoint variant, resulting in a 15.7-point R@1
advantage over NU at Window 4 (75.6 vs.\ 59.9) with the same change
detection pipeline.

\textbf{R+V occupies the efficient corner.} As shown by the Window~4
endpoints in Fig.~\ref{fig:vpr_scatter}, R+V achieves higher R@1 than NA at Window 4
(75.6 vs.\ 74.8) while maintaining a database 27\% smaller (26,829 vs.\
36,543 images). NA and A require $\sim$9.7k more images than R+V to achieve
lower recall, confirming that pool inflation from indiscriminate appending
degrades retrieval precision. The R strategy, which skips viewpoint pairs,
achieves only 70.6 R@1 at constant size, demonstrating that viewpoint
coverage augmentation (the +V component) is essential and not merely
redundant with structural replacement.

R@1 is the operationally relevant metric for single-shot localization, where a robot acts on only the top retrieved match, and we highlight it accordingly in Fig.~\ref{fig:vpr_results}. Table~\ref{tab:extended_recall} reports R@5/10/20 for all strategies and windows from the same evaluation runs; the ranking established under R@1 holds across all four recall depths.

\textbf{Descriptor generality.} The experiments above use
MixVPR~\cite{ali2023mixvpr} for retrieval, consistent with its use in
constructing NYC-CD itself. The update mechanism, however, is
descriptor-agnostic: it requires only a top-1 visual retrieval step
(Sec.~\ref{sec:vpr_maintenance}) and does not depend on any property specific
to MixVPR's embedding space. We therefore expect the strategy ranking
observed here (R+V $>$ NU $>$ A $\approx$ NA $>$ R) to hold under alternative
descriptors such as SALAD~\cite{izquierdo2024optimal}, since the ranking is
driven by \emph{which} images are selected for update, not by the retrieval
backbone that supplies candidates; we leave empirical confirmation to future
work.

\textbf{Geometric verification failure modes.} MAST3R's dense matching is
reliable under the moderate viewpoint and seasonal appearance shifts observed
across our four windows, but its behavior under more extreme conditions is
not separately characterized here. Under near-180\textdegree{} viewpoint
reversals or drastic illumination change, dense correspondence estimation may degrade, either rejecting
genuine same-location pairs or passing perceptual aliasing between repeated urban structures.
We do not observe evidence of the latter failure mode within our evaluated
windows, but robustness under such extreme conditions remains an open
question for deployment in less constrained settings.

\section{CONCLUSIONS}
\label{sec:conclusion}

SCD4VPR demonstrates two findings for long-term VPR database
maintenance. First, retrieval recall deteriorates substantially when
a database is left unchanged as the environment evolves,
with recall dropping by tens of points at the largest observed time
gap. Second, multimodal SCD can recover
most of this lost recall while keeping the database compact, rather
than trading recall for unbounded growth as naive append-based
maintenance does.

Underlying this result is a finding largely overlooked by prior SCD
work: viewpoint-induced change, rather than being discarded as noise,
is itself a valuable signal for VPR maintenance. Naive replacement
of viewpoint-change-only pairs
cumulatively degrades retrieval, whereas \textbf{R+V}'s strategy of \textit{preserving} the original entry and \textit{adding} the new viewpoint variant avoids this degradation, which cannot be achieved by binary change detections, as it cannot distinguish a viewpoint shift from a structural change.

\textbf{Limitations.} Our database update strategies are simple, fixed policies; whether learned or budget-constrained (e.g., update-capped) curators could improve further on R+V's recall-per-added-image efficiency is left to future work. SCD4VPR also currently operates offline at 4.89\,s per pair; replacing the online VLM call with a lightweight cached or distilled language model may enable faster online operations.

Beyond these limitations, we believe the core idea of preserving valid coverage rather than overwriting it could generalize beyond VPR to other retrieval system whose database degrades under viewpoint drift, including visual SLAM loop closure and augmented reality relocalization.






\section*{ACKNOWLEDGMENT}
This work is supported in part by NSF Grant \#2238968 and \#2345139. We thank
the NYU HPC team for their assistance and support.


\makeatletter
\patchcmd{\thebibliography}{\footnotesize}{\scriptsize}{}{}
\makeatother
 
\balance
\bibliographystyle{IEEEtran}
\bibliography{IEEEabrv,references}

\end{document}